# BIQ2021: a large-scale blind image quality assessment database

**Nisar Ahmed**[a,*] **and Shahzad Asif**[b]
[a]University of Engineering and Technology Lahore, Department of Computer Engineering, Lahore, Pakistan
[b]University of Engineering and Technology Lahore, Department of Computer Science, Lahore, Pakistan

**Abstract.** Perceptual quality assessment of digital images is becoming increasingly important due to widespread use of digital multimedia devices. Smartphones and high-speed internet are among the technologies that have increased the amount of multimedia content by several folds. Availability of a representative dataset, required for objective quality assessment training, is therefore an important challenge. We present a blind image quality assessment database (BIQ2021). The dataset addresses the challenge of representative images for no-reference image quality assessment by selecting images with naturally occurring distortions and reliable labeling. The dataset contains three set of images: images captured without intention of their use in image quality assessment, images obtained with intentional introduced natural distortions, and images collected from an open-source image sharing platform. Ensuring that the database contains a mix of images from different devices, containing different type of objects, and having varying degree of foreground and background information has been tried. The subjective scoring of these images is carried out in a laboratory environment through single-stimulus method to obtain reliable scores. The database provides details of subjective scoring, statistics of the human subjects, and the standard deviation of each image. The mean opinion scores (MOSs) provided with the dataset make it useful for assessment of visual quality. Moreover, existing blind image quality assessment approaches are tested on the proposed database, and the scores are analyzed using Pearson and Spearman's correlation coefficients. The image database and the MOS along with relevant statistics are freely available for use and benchmarking. *© 2022 SPIE and IS&T* [DOI: 10.1117/1.JEI.31.5.053010]

## 1 Introduction

Proliferation of multimedia content has increased over the past decade due to rapid development of digital multimedia devices. Smartphones with cheap and good quality image acquisition and processing hardware have made it more common. Availability of high-speed internet and social media platforms has increased the creation and sharing of multimedia content. The quality of these images is a major concern because the low-quality images are less useful, less appealing, and sometimes annoying.[1,2] Moreover, creation of multimedia content by novice user results in low-quality content containing blur, noise, over- and under-exposure, and other quality issues.[3] Because humans are the ultimate consumer of these images, perceptual quality of these images is important. To ensure the provision of high-quality images, objective quality metric with adequate accuracy is required. Subjective quality assessment is the most reliable for quality assessment of digital images but it is expensive, time-consuming, and difficult to use in many scenarios. Objective quality assessment methods therefore design metrics that attempt to adequately model the behavior of human visual system (HVS). Despite various efforts to design objective quality metrics, none has provided reliable modeling HVS, capturing all the peculiarities. However, there are several limitations in the design and use of these metrics.[4–8] A limitation with these systems is the understanding of the HVS as it is a complex system, and therefore it is difficult to study and model HVS. Therefore, existing metrics try to incorporate some specific features that affect the visual perception. In contrast, deep learning methods try to learn these features on their own, and therefore a representative dataset of sufficient size and quality is required to build a successful model.

*Address all correspondence to Nisar Ahmed, nisarahmedrana@yahoo.com

Creation of a representative dataset with a good number of images having reliable quality scores is difficult in aspects such as collection of representative images and obtaining mean opinion scores (MOSs) based on subjective test. Many human subjects are required to evaluate a large number of images, which is an expensive operation both in collection of images and acquisition of MOS. The MOS is a particularly important parameter because the supervised training will be guided by this and only a reliable MOS will provide a model, which can truly represent human judgment. Full-reference quality databases have a lot of choices to perform subjective experiments due to availability of reference and distorted images but the no-reference database has only one image, and absolute quality rating is considered a standard method for multimedia applications.[9] According to ITU-T P.910[9] recommendations, absolute category rating with fixed number of discrete scales should be used for subjective evaluation. The subjective evaluation is required to be performed by at least 10 volunteers, whereas more evaluation from more than 30 volunteer with diverse background will provide a reliable judgment. The MOS will be obtained by averaging the judgment of these volunteers, whereas variations in MOS have been developed for some specific domains.[10,11] It is required to perform the subjective experiment in a laboratory environment with suitable hardware and display device. The set of images and interface for subjective experiment as well as the instructions should be the same for each volunteer. The reliability and repeatability of these experiments and obtained MOS will strongly depend on the degree of conformation to the stated requirements.

The reasons for creating a new database are to improve the existing databases and provide a benchmark for testing of a trained model on images having naturally occurring distortions. Most of the existing databases are intended for evaluation of full-reference image quality assessment (IQA) tasks and contain images with simulated distortions. Most of the datasets contain single distortion per image,[12–17] whereas some datasets provide images with multiple simulated distortions per image.[18] These databases can be used for evaluation full-reference approaches or the scenario where the type of distortion is known,[19–22] i.e., performance evaluation of a compression algorithm. The no-reference models trained on these databases fail to generalize in the scenarios where the distortion space is multidimensional. A database with naturally occurring distortions is required to model the problem where multiple concurrent distortions are in play.

The necessity of the new database, blind IQA database (BIQ2021), is justified based on the following justifications:

1. Intensive use of existing databases reveals their limitations mainly due to limited number of simulated distortions. A model with enough training parameter can easily learn the simulated distortion behavior, which has specific discrete levels (five in most cases) of distortions and a fixed number of distortion types.
2. A lot of existing metrics have either simulated distortions or a few images in case of real distortions. Moreover, the newer metrics have learned their peculiarities and provided better performance, whereas their generalization power is not improved to a reliable level. Moreover, many new datasets with a large number of images with naturally occurring distortions have employed online crowdsourcing, whereas the perceptual quality of images greatly varies with size type of display, viewing angle and distance, ambient effect, and other parameters. Subjective scores obtained in crowdsourcing experiments are therefore less reliable unless extensive measures are taken to screen the subjects and viewing conditions.
3. This database will create a competition in the research community by creating a new benchmark for comparison. Feature maps driven no-reference image quality evaluator engine (FRIQUEE)[23] provided the best score for Live in the Wild challenge database, from the same author,[3] in 2015. Whereas, in the next few years, many approaches have significantly surpassed the claimed performance. This type of competition is positive and will spark the research community to explore new methods for the design for visual quality metrics.

4. The evaluation of technology has provided new consumer electronic devices, such as smartphones and cheap digital cameras, that are characterized by new type and combination of distortions. Test sets of images captured with such devices are therefore required to assess the quality assessment performance of the models.

## 2 Related Work

Training and evaluation of an IQA algorithm requires a database of images with quality score for each image. The quality score is usually obtained through subjective scoring and provided in the form of MOS. A number of databases for IQA have been proposed for design and evaluation of full- and no-reference IQA algorithms. The databases designed for full-reference algorithms contain reference images along with their corresponding distorted images produced by simulation. These databases can be used for no-reference quality assessment by discarding the reference images. In contrast, the no-reference IQA databases contain naturally occurring distortions rather than simulated distortions and cannot be used for development of full-reference models. Table 1 provides some of these databases separated into databases with simulated and natural distortions.

Some of the most commonly used databases for IQA are image and video communication (IVC),[24] laboratory for image and video engineering (LIVE),[25] tampere image databases (TID2008[26] and TID2013[27]), and categorical subjective image quality (CSIQ).[28] Some other less known conventional IQA databases are multimedia information and communication technology (MICT),[29] multiply distorted image database (MDID),[30] multiply distorted image quality, LIVE multiply distorted (LIVE MD),[31] and blur image database (BID).[32] The databases are created by collecting few images of pristine image quality and then distorted through a simulation model. Multiple levels of degradation are used to distort a single image in most of the scenarios. Some of the image databases pass the image through multiple simulation stages to reflect the fact that natural images face several simultaneous distortion. These databases can be used for full- and no-reference IQA because each image has its corresponding reference image. However, they cannot truly represent the case of natural distortion, which occurs during the process of image acquisition, processing, or storage.

Konstanz artificially distorted image quality database (KADID-10k),[17] Konstanz artificially distorted image quality set (KADIS-700k),[17] and Waterloo exploration[33] database are the recent image quality databases. The above-mentioned databases are designed to cover the limitations of small-scaled databases used conventionally. These large-scale databases can benefit the development of deep-learning-based IQA methods. KADID-10k contains 81 pristine images each distorted by 25 distortions in 5 distortion levels. The database contains a total of 10,125 distorted images can be used for no-reference quality assessment. The subjective study is conducted through the use of crowdsourcing, and 30 degradation category ratings are obtained per image. In contrast, KADIS-700k[17] contains a total of 140,000 pristine images with randomly chosen distortion categories and 5 distortion levels. The database generates a total of 700,000 distorted images without quality scores. Waterloo exploration database contains 4744 pristine images and 94,880 corresponding distorted images. Unlike other database, they have not provided quality scores and introduced another alternative quality evaluation mechanism.

The problem of small-scaled database is somehow resolved by KADID-10k and Waterloo exploration databases but the originality of the distortions still remains an issue. Most of these databases generate distorted images by simulating discrete distortion levels of specific interference. In real scenario, multiple distortions occur in varying degree simultaneously that is challenging to model. These databases can be useful in limited distortion scenarios such as compression, Gaussian noise, motion blur and etc. Therefore, image databases with synthetic distortion are not suitable candidates for general-purpose IQA. Moreover, content diversity of these databases is also limited. CID2013[34] was the first database in our knowledge to introduce natural distortions. They created the database having 480 images captured in 8 different scenes with 79 image capturing devices. Their image acquisition method is time-consuming and expensive and thus not suitable for large-scale database generation.

**Table 1** Summery of IQA databases.

| Database | Year | Images | Reference images | Distortion types | Simultaneous distortions | Ratings/image | Image resolution | Subject expertise | Method |
|---|---|---|---|---|---|---|---|---|---|
| | | | | | **Simulated distortions** | | | | |
| IVC[24] | 2005 | 185 | 10 | 4 | 1 | 15 | 512 × 512 | Expert | Laboratory |
| LIVE[25] | 2006 | 779 | 29 | 5 | 1 | 23 | 768 × 512 | Naive | Laboratory |
| MICT[29] | 2008 | 196 | 14 | 2 | 1 | 16 | 768 × 512 | Naive | Laboratory |
| TID2008[26] | 2009 | 1700 | 25 | 17 | 1 | 33 | 512 × 384 | — | Laboratory |
| CSIQ[28] | 2009 | 900 | 30 | 6 | 1 | 5 to 7 | 512 × 512 | — | Laboratory |
| TID2013[27] | 2013 | 3000 | 25 | 24 | 1 | 9 | 512 × 384 | Naive | Laboratory |
| BID[32] | 2011 | 585 | | 5 | 2+ | 11 | Variable | — | — |
| LIVE MD[31] | 2012 | 405 | 15 | 3 | 2 | 18 to 19 | 1280 × 720 | Naive | Laboratory |
| Waterloo exploration[33] | 2016 | 94,880 | 4774 | 4 | — | 30+ | 512 × 512 | — | Laboratory |
| MDID[30] | 2017 | 1600 | 20 | 5 | 4 | 33 to 35 | 512 × 384 | Naive | Laboratory |
| KADID-10k[17] | 2019 | 10,125 | 81 | 25 | 1 | 30 | 512 × 512 | Naive | Crowdsourcing |
| KADIS-700k[17] | 2019 | 700,000 | 140,000 | 0 | 1 | 0 | 512 × 512 | — | NA |
| | | | | | **Natural distortions** | | | | |
| CID2013[34] | 2013 | 474 | NA | NA | — | 31 | 512 × 512 | Naive | Laboratory |
| LIVE in the Wild[3] | 2016 | 1169 | NA | NA | — | 175 | 512 × 512 | Naive | Crowdsourcing |
| KonIQ-10k[11] | 2018 | 10,073 | NA | NA | — | 120 | 512 × 384 | Naive | Crowdsourcing |
| PaQ-2-PiQ[35] | 2020 | 39,810 | NA | NA | — | ~33 | 640 × 640 | Naive | Crowdsourcing |
| SPAQ[36] | 2020 | 11,125 | NA | NA | — | 600 | 512 × 512 | Naive | Laboratory |
| HTID[37] | 2021 | 10,073 | NA | NA | — | 120 | 512 × 384 | Naive | Crowdsourcing |
| BIQ2021 | 2020 | 8000 | NA | NA | — | 20+ | 512 × 512 | — | Laboratory |

Live in the Wild[3] is another natural distortion image database having 1169 images captured by different mobile cameras. They have used crowdsourcing experiments for subjective scoring. Although they provided relatively reliable scoring and larger number of images, the content diversity and database sizes are small. The crowdsourcing experiments have reduced the time of subjective scoring at cost of reliability. KonIQ-10k[11] is a recent database having natural distortions. The image database contains 10,073 naturally distorted images with diverse content and subjective scores obtained through crowdsourcing experiments. To date, the database is largest natural distortion database having diverse content but relatively less reliable quality score due to crowdsourcing experiments.

Table 1 provides a summary of 12 simulated distortions database and 4 natural distortion databases.

## 3 Database Creation

The problem of blind IQA (BIQA) is traditionally addressed by collecting a set of pristine images and then generating a set of distorted images through simulation. The distortion mechanism is modeled to simulate these distortions. These images are used for subjective scoring using either single- or multi-stimulus approaches. The MOS is calculated by arithmetic mean of the subjective scores obtained from various experiments. It is important to note that there are several approaches for performing subjective experiments or obtaining the target score (i.e., MOS), each with their own set of advantages and disadvantages. Most of these image databases do not contain multiple distortions per image and thus are not true representative of naturally occurring distortions. Furthermore, naturally occurring images suffer from noise, blur, compression, and other distortions at the same time, which these image databases do not. These databases contain only a small number of distortion types with discrete distortion scales, and it is possible that the learning algorithm will learn the peculiarities of the simulation system rather than the behavior of HVS.

In the case of natural distortions, such as CID2013, Live in the Wild challenge database, and KonIQ-10k, CID2013 contains relatively smaller number of images, whereas the other two databases contain a larger number of images but the scoring is obtained through less reliable online crowdsourcing experiments. To address these limitations, we have provided a database with the following characteristics: The database contains images that were distorted during the acquisition, processing, and storage processes rather than during simulation experiments. It provides several simultaneous distortions in images that occur naturally, as well as a true representation of BIQA modeling.

To avoid aesthetic bias during subjective evaluation, the database contains images with a variety of content. Although it is nearly impossible to overcome this bias due to human psychology, a greater diversity of content and a greater number of human subjects from diverse backgrounds have helped to minimize this bias. We have gathered 30 ratings per image because subjective scoring is an expensive method and conducting it in a laboratory environment under supervision is even more challenging. Therefore, to meet the requirements of reliable ratings, we have collected these ratings that are just enough to meet the requirements.[9]

The database contains three image subsets, each of which serves the purpose of providing a diverse set of images. The first subset contains 2000 images chosen from an image gallery of images captured by N. Ahmed between 2007 and 2020. These images have varying degrees and types of distortions captured by various image acquisition devices. These images have distortions due to camera and photographic error, as well as distortions introduced during processing and storage. Because these images were not captured for the purpose of IQA, they can serve as a true representative for evaluating IQA algorithms. Figure 1 shows some randomly selected images from the BIQ2021 database's subset-1. The second subset contains images that were taken with the intention of being used for IQA. This subset contains 2000 images, and it ensures that the entire spectrum of quality scoring is covered by introducing images ranging from the worst to the best. The distortions in this subset were intentionally introduced during the acquisition process, similar to CID2013, but we did not use fixed levels of distortion. The images are captured by changing the International Organization for Standardization (ISO) from 50 to 3200 and the shutter speed from 4 to 1/1250 s. Auto- and manual focuses are used for acquisition, and lens blur and motion blur are introduced on purpose during acquisition. To introduce the effect of ambient light when photographing an indoor environment, the ambient light is changed. Because there are very few images with extreme distortion levels, this subset was created to balance different distortion levels. Figure 2 shows images from the BIQ2021 database's subset-2.

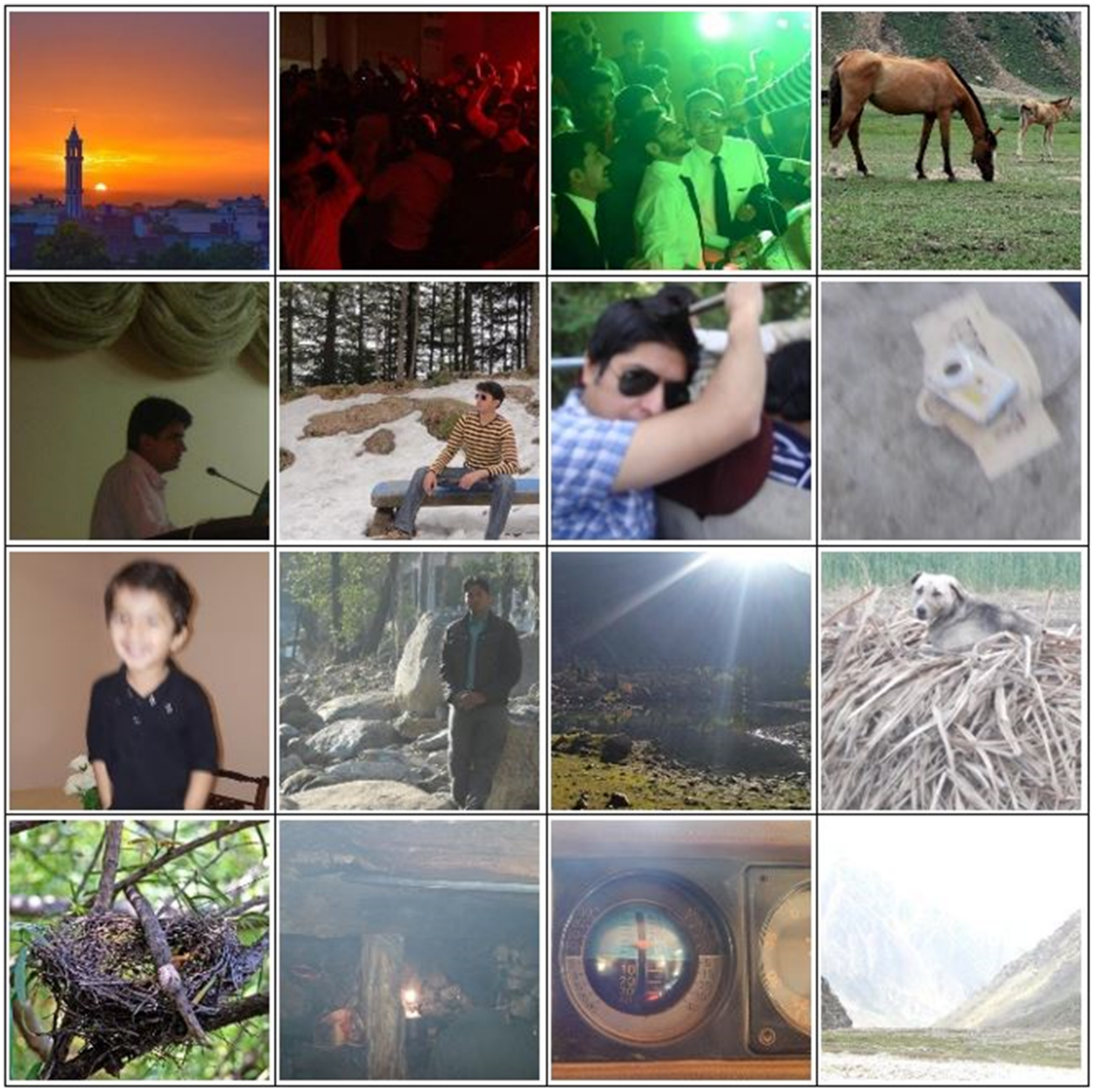

**Fig. 1** Sixteen sample images from subset-1 of BIQ2021.

The third subset of 8000 images is acquired from Unsplash.com, where use of images for scientific or commercial purposes is allowed. These downloaded images are searched for using various keywords to introduce diversity of content and are specifically chosen for the purpose of IQA. The keywords used for search are animals, wildlife, pets, birds, zoo, vegetables, fruits, food, cooking, architecture, cityscape, night, indoor, outdoor, scenery, mountain, lake, candid, close-up, experimental, texture, people, men, women, model, kids, babies, boy, girl, fashion, culture, vintage, sports, and swimming. It should be noted that many of these images have been postprocessed, making them an excellent candidate for learning the effect of postprocessing on perceptual quality. Furthermore, because it contains images with a wide range of content, this subset of images contributes to the database's diversity. Figure 3 shows some selected images from this category.

### 3.1 *Image Cropping*

Image size normalization is important because image scale affects perceptual quality.[38] As a result, we cropped and resized the images to provide a consistent image size. Image resizing is performed using bicubic interpolation due to its satisfactory performance and computational complexity. The images are all true color, 24-bit images with $512 \times 512$ image dimensions. Most images are not square in shape, and shrinking them will change their aspect ratio, which will affect the perceived image quality.[39] As a result, we cropped these images to the image's smaller dimension, resulting in a square image that is then resized to standard dimensions. The crop location is determined by the visual saliency detection method presented in Ref. 40, which is quick and serves our intended purpose well.

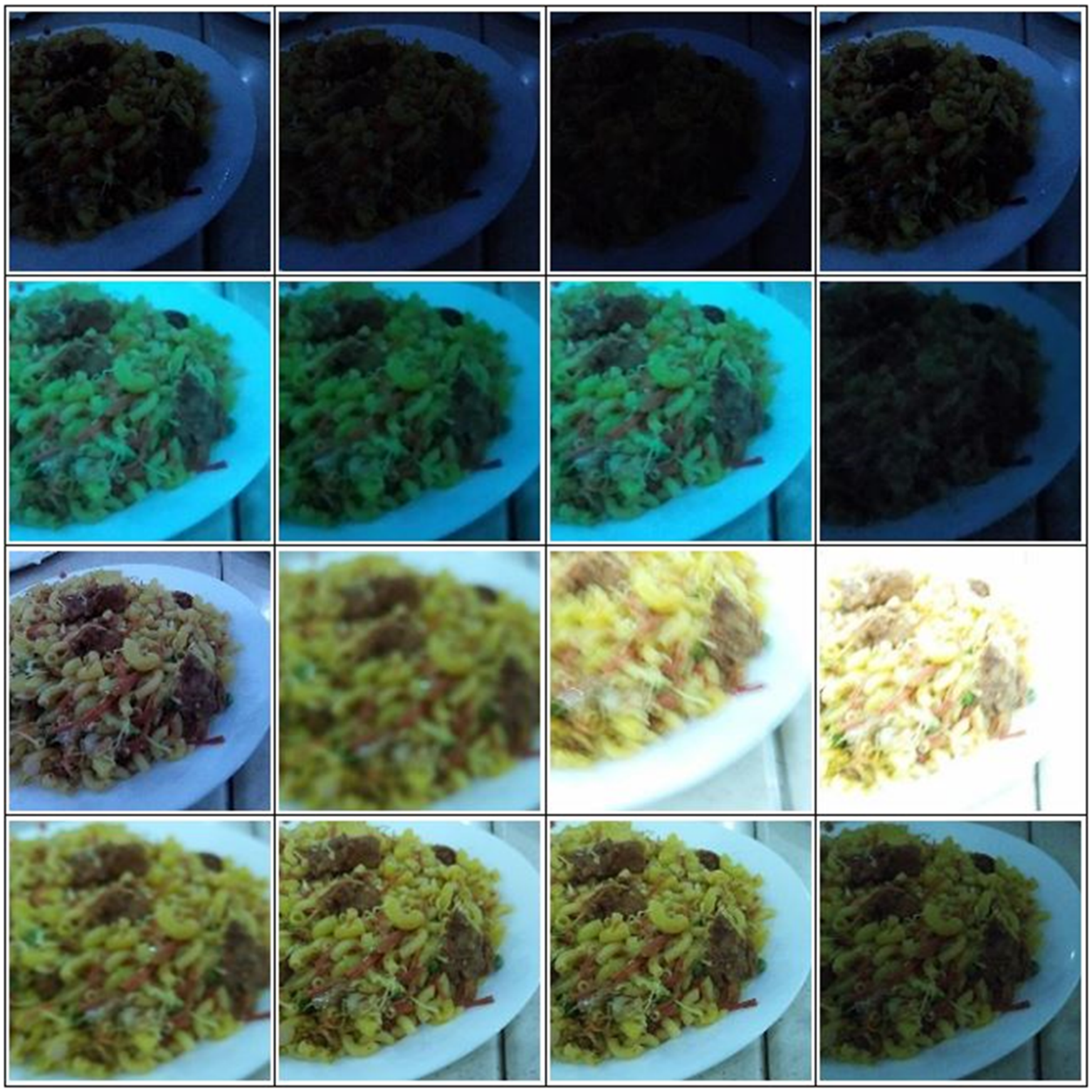

**Fig. 2** Sixteen sample images from subset-2 of BIQ2021.

## 4 Subjective Scoring Experiments

Subjective quality scoring experiments are carried out to obtain the MOS. In this paper, various approaches are used, each with its own set of advantages. Subjective quality scoring experiments are broadly divided into two types: laboratory-based (supervised)[12,34] and crowd-sourcing (online).[11,17] The credibility of crowdsourced subjective quality experiments is being questioned, despite the fact that numerous measures have been implemented to improve reliability.[3,11,17] Laboratory experiments are typically carried out under the supervision of a tutor who guides the participant throughout the experiment.[12] The experimental conditions, such as display, ambiance, viewing distance, and angle, are monitored and are thus considered more reliable.

The scoring is obtained from the user through either a single- or a multi-stimulus approach. A single stimulus provides only one image for scoring at a time and requires an absolute score. The multi-stimulus method displays two or more images, one of which is typically the reference image. The subject assigns a relative score to images, which is more convenient, but this method is only applicable to full-reference scoring methods. Although more exhausting, the single-stimulus method is the suitable method for no-reference scoring.

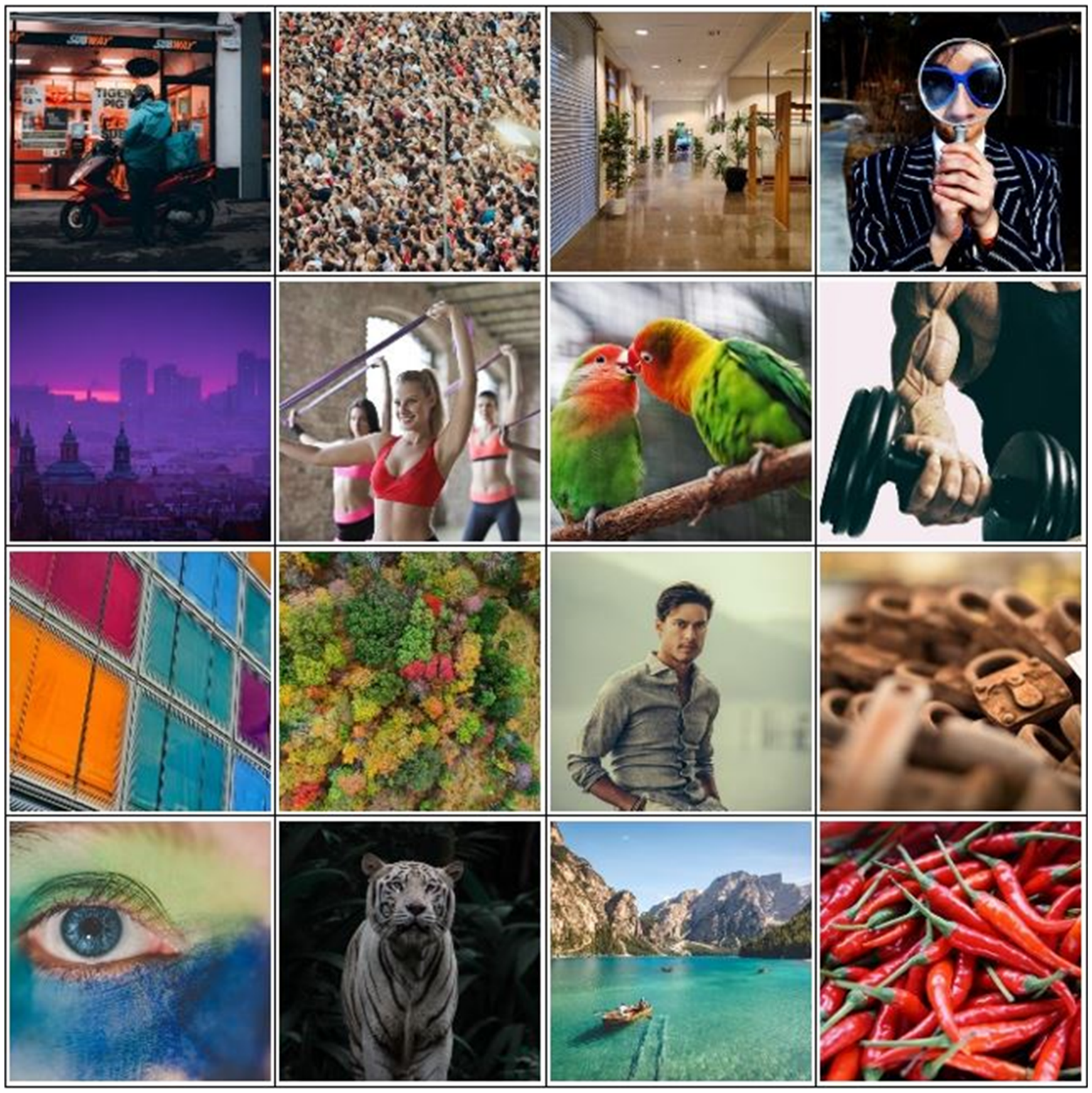

**Fig. 3** Some selected images from subset-3 of the database.

To obtain a representative subjective score, a scoring mechanism or scale is required. Some of the methods used for this purpose include a continuous or discrete scoring scale, categorical or numerical inputs, and pair-wise or relative image comparison. The pairwise comparison method is thought to be the most user-friendly for subjects.[13] Due to the lack of a reference image, absolute category rating is the appropriate approach for blind quality scoring. In the case of categorical scoring, the user is given three or five options for rating the quality of an image. The following are some examples of five-grade categorical scoring: excellent, good, fair, poor, and bad. When it comes to numerical scales, they can be continuous in the form of a sliding bar or discrete with 5 to 10 levels. We chose the five quality level scale with distinct categories.

## 4.1 *Experimental Settings*

The graphical user interface (GUI) for the experiment is designed in MATLAB® 2019b with image displayed on left side and the slider provided under the image. The absolute score assigned based on the slider position is also provided in a textbox. A line graph of their score history is also provided on the right side to guide the user in recalling their previous answers and encouraging them to use the full scale. Figure 4 shows the GUI used during the experiment, depicting the displayed image on left, slider and text box for quality score assignment, and line graph of score history on right.

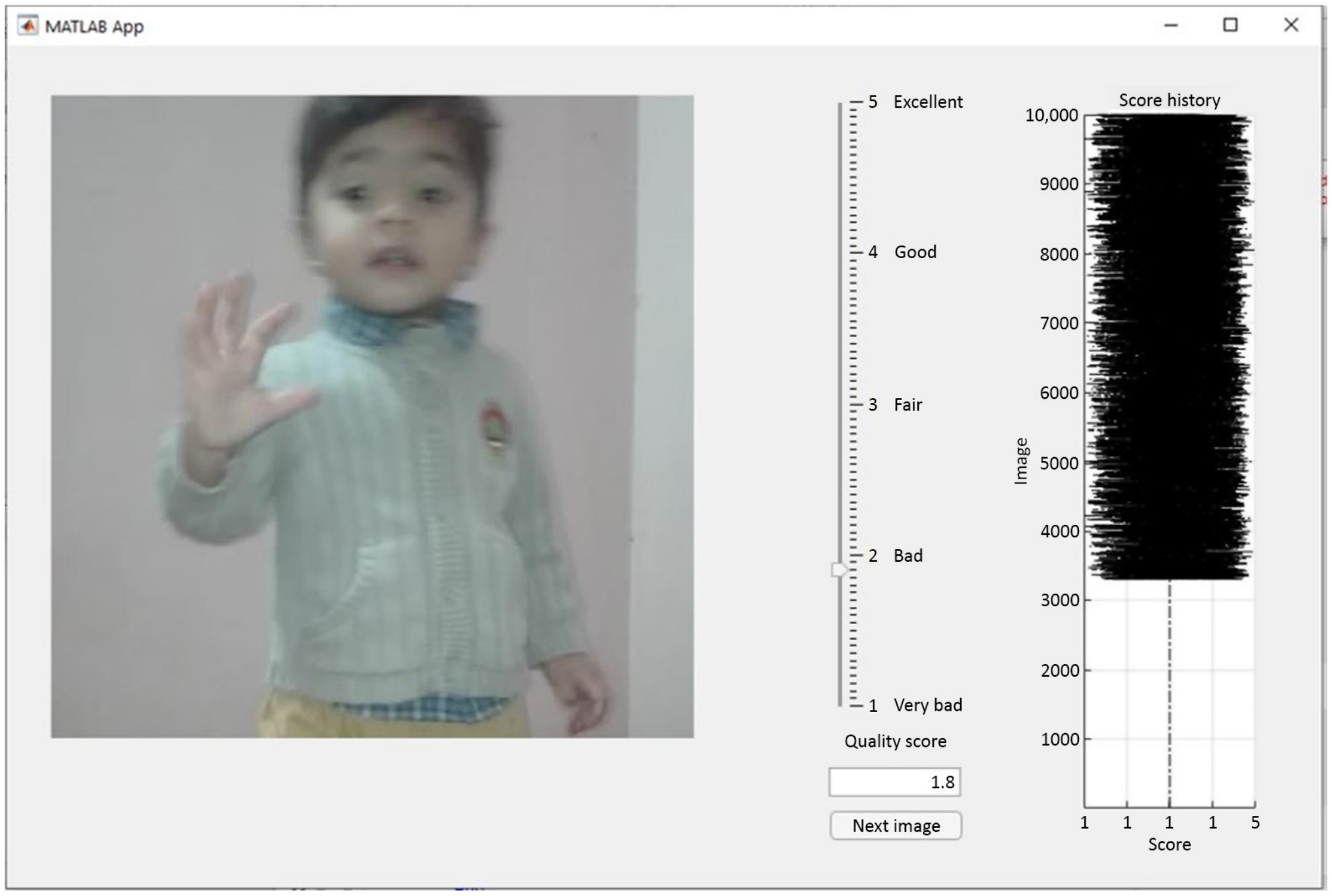

**Fig. 4** GUI used for subjective scoring experiments.

The experiment was carried out in an indoor setting with diffused lighting and a comfortable temperature. Participants were instructed to sit within 60 in. of a 24-in. liquid-crystal display. However, the subjects were allowed to adjust the distance or chair height to their comfort level. The experiment began with instructions regarding project guidelines, the purpose of the subjective experiment, and the scoring criteria. Every subject was either a researcher or a graduate student. Three of the subjects wore glasses to correct their vision. In the IQA study, all of the subjects were naive. The subjects were allowed to take a break and continue the test if they were bored or tired because boredom and tiredness would have a negative impact on the reliability of the quality scores. The subjective quality scoring experiment lasted 4 months, and 30 subjects provided a total of 0.36 million ratings.

## 4.2 *Score and Label Computation*

The scores assigned by the subjects to images contained five distinct levels. The distinct quality levels with tag, such as bad or very bad, are more realistic and allow the participant to provide reliable labels. We have therefore obtained two types of labels from these scores. The one is category label, which allows the researcher to train the IQA as a classification problem. The other is a MOS, which will be used for conventional regression-based approach. The MOS is obtained by taking arithmetic mean of the quality scores assigned by 30 participants. The resulting MOS is rescaled using min–max scaling to a range of 0 to 1

$$S_i' = \frac{S_i - \min(S_i)}{\max(S_i) - \min(S_i)}. \tag{1}$$

The categorical class labels are computed using majority-vote classifier approach popularly used in hard voting ensembles. This approach works by counting the occurrence of each class label and then outputs the class label that is assigned by maximum participants. It is possible that a participant may assign an incorrect class label to an image but the combined effect of 30 participants is very less likely to contain an incorrect class label. The equation for majority vote classifier is given as

$$C(x) = \arg\max_i \sum_{j=1}^{P} x_i I(x_j(X) = i), \tag{2}$$

where $P = 30$ and $w_i$ is the weight that is chosen $1/P = 1/30$ because all the participants are given equal weightage. $h_j(X)$ is the class label assigned by participant $j$ and $I$ in the indicator function.

The count of images falling in each quality category is provided in Table 2 with highest number of images falling in category good and lowest number of images falling in category excellent. The histogram of MOS is provided in Fig. 5 indicating that quality scores assigned to each image are spread over the complete spectrum. It can also be noted that the distribution is quite symmetrical and centered at 60% quality. The standard deviation for each image is also calculated and provided along with the scores. The histogram of the standard deviation is provided in Fig. 6.

**Table 2** Count of images in each quality category.

| Quality label | Count of images |
|---|---|
| Excellent | 714 |
| Good | 3759 |
| Fair | 2927 |
| Bad | 2854 |
| Very bad | 1746 |

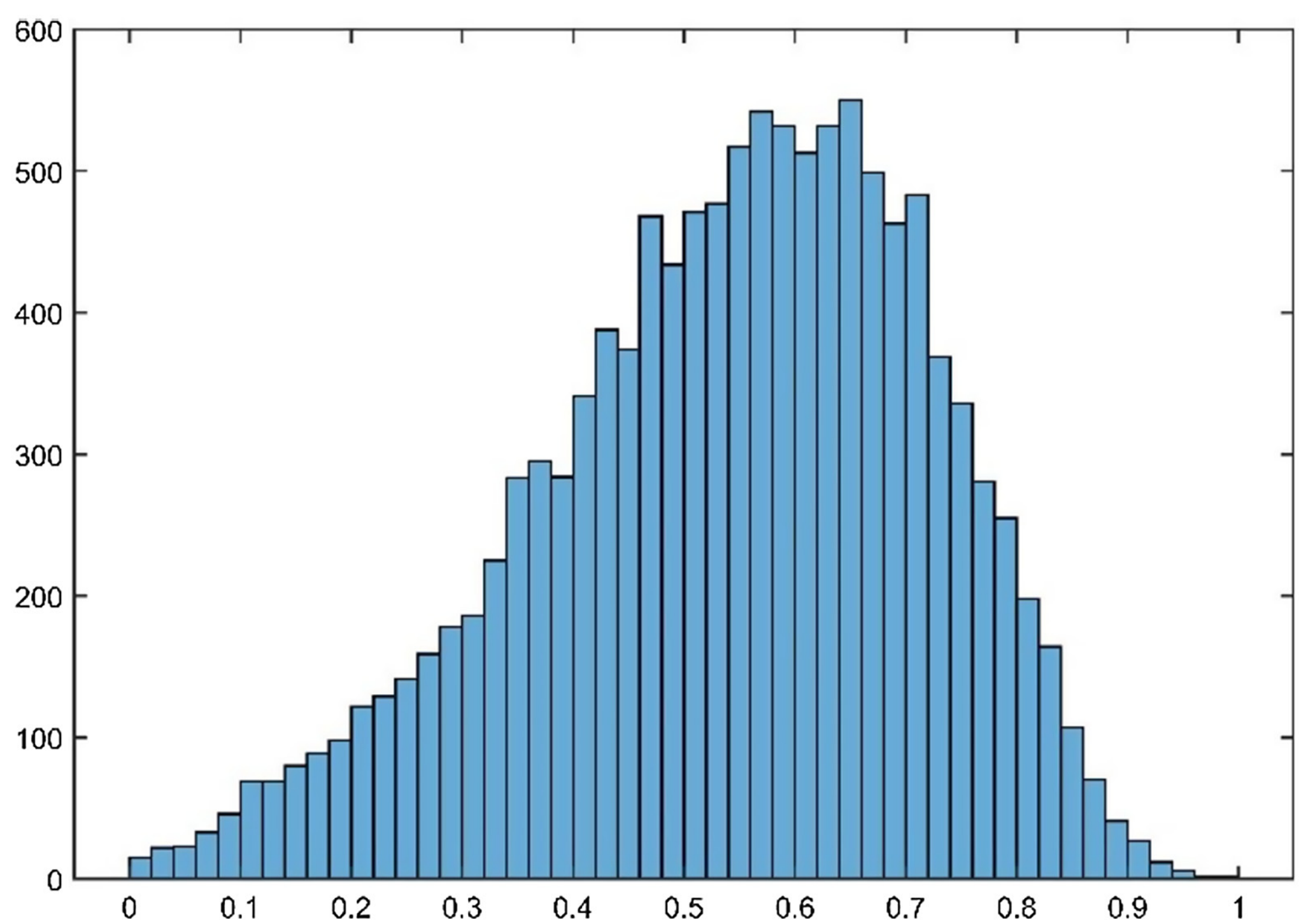

**Fig. 5** Histogram of MOS with Gaussian fitting.

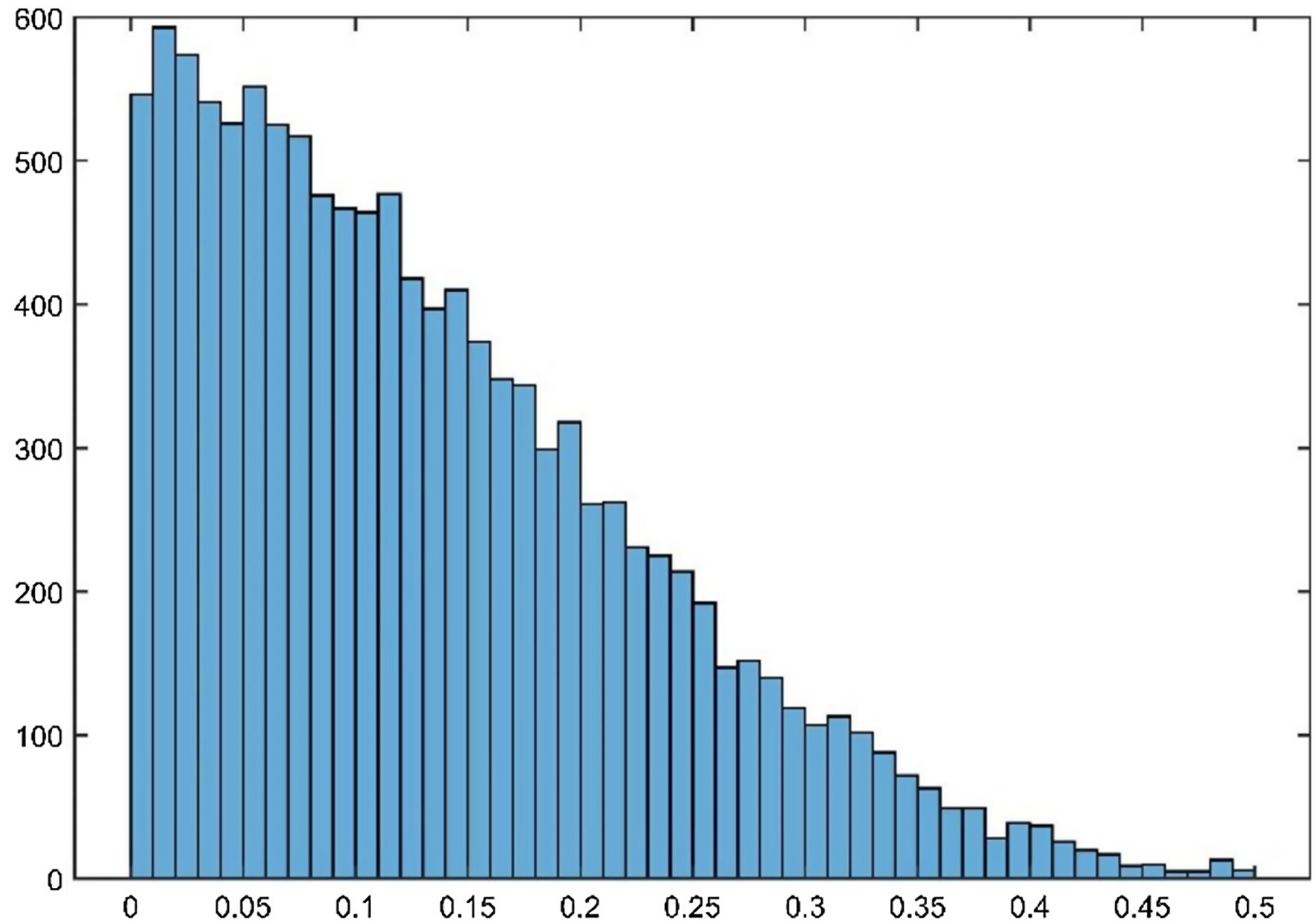

**Fig. 6** Histogram of standard deviation for images.

## 5 Performance Evaluation and Discussion

### 5.1 *Classification Performance of BIQ2021*

BIQ2021 was developed to serve as an alternative benchmark for evaluating the performance of BIQA methods. It is one of the few image databases with naturally occurring distortions and is intended to be used as a suitable candidate for evaluating existing and new blind IQA methods. We trained several deep learning architectures with a simple modification to train as an image quality evaluator. It is proposed to train an image quality evaluator with discrete quality labels because they are more descriptive, easy to understand, and are governed by strong loss functions when modeling a classification problem.

Furthermore, humans are less concerned with an image's absolute quality score because they prefer quality labels that indicate whether the image is of good or poor quality. As a result, the predicted absolute quality score is eventually converted to a binary or multilevel decision function, making the quality category a better option for image quality evaluators. Figure 7 shows the classification network architecture, in which the convolutional neural networks (CNN) base is replaced by each architecture under test (Fig. 7).

The results of Table 3 show that NASNet has the highest prediction accuracy among 13 models developed from popular CNN architectures. The reason behind high prediction performance of NASNet is its better and deeper architecture.

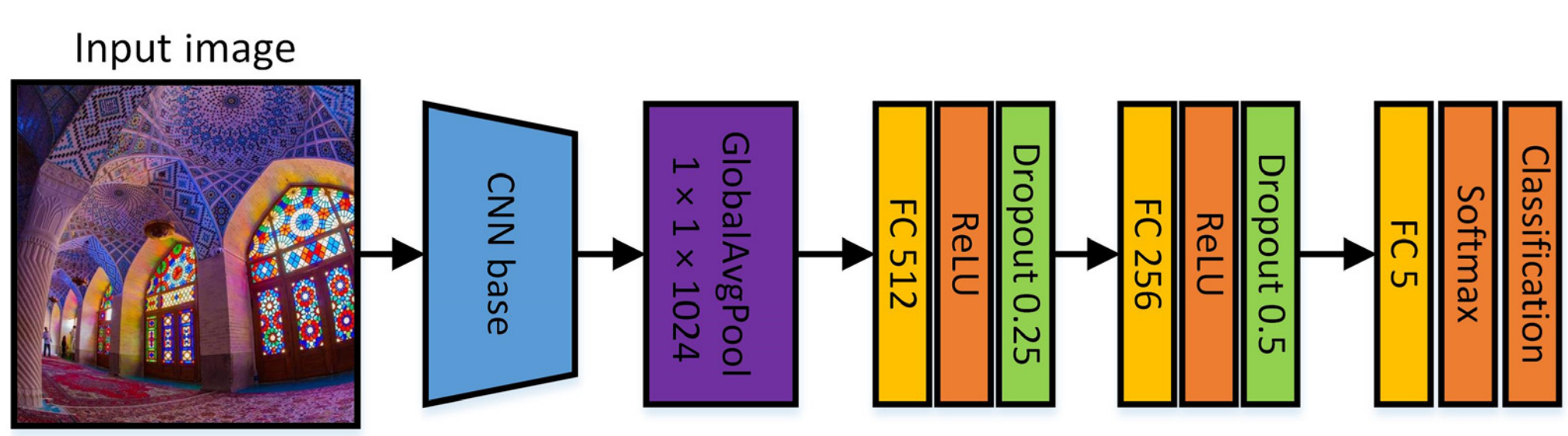

**Fig. 7** Architecture of CNN-based classification model.

**Table 3** Performance of CNN architectures used for classification modeling.

| Number | Architecture | Parameters (millions) | Depth (layers) | Accuracy (%) |
|---|---|---|---|---|
| 1 | AlexNet | 61.0 | 8 | 74.09 |
| 2 | DarkNet-53 | 41.6 | 53 | 79.78 |
| 3 | DenseNet201 | 20.0 | 201 | 85.95 |
| 4 | EfficientNet-B0 | 5.3 | 82 | 80.86 |
| 5 | Inception-ResNet-V2 | 55.9 | 164 | 85.98 |
| 6 | Inception-V3 | 23.9 | 48 | 82.15 |
| 7 | MobileNet-V2 | 3.5 | 53 | 81.93 |
| 8 | NASNet | 88.9 | + | 90.30 |
| 9 | NASNet-Mobile | 5.3 | + | 79.45 |
| 10 | ResNet101 | 41.6 | 101 | 85.75 |
| 11 | ResNet50 | 25.6 | 50 | 86.02 |
| 12 | Vgg16 | 138.0 | 16 | 83.55 |
| 13 | Xception | 22.9 | 71 | 80.59 |

## 5.2 *Regression Performance of BIQ2021*

The most common annotation for image quality databases is MOS having continuous quality scores. As a result, the majority of image quality evaluators is regression-based and attempt to predict the absolute quality value for each image. Figure 8 shows the CNN architecture used for regression, which is mostly similar to that of a classification network except for the last few layers. Similarly, to the classification model, the CNN base is replaced with the CNN model under test's base architecture. The results of Table 4 show that NASNet provided the best regression performance in the form of Pearson Linear Correlation Coefficient (PLCC) and Spearman's Rank Order Correlation Coefficient (SROCC).

## 5.3 *Training Options*

The CNN architectures are trained using the Adam optimizer with 8, 16, or 32 batch size that can be loaded on the GPU. The maximum number of epochs is set to 300, with a validation patience of three checks, which means that the training will be terminated if the validation performance does not improve after three validation checks. The initial learning rate is set to $3(10)(-3)$, and the image sequence is shuffled after each epoch.

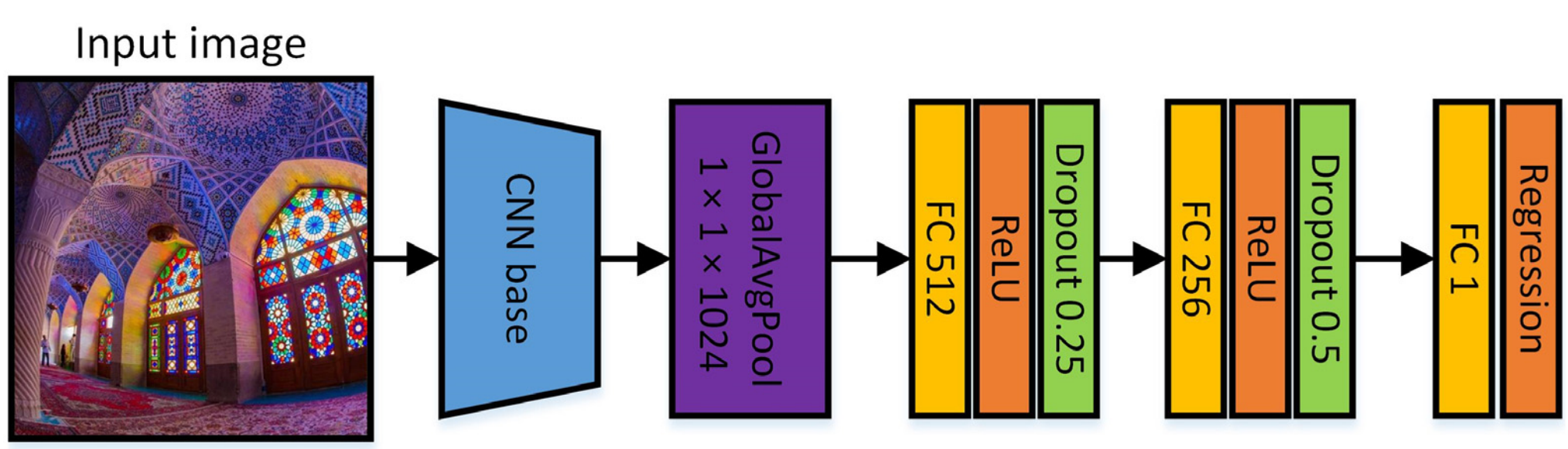

**Fig. 8** Architecture of CNN-based regression model.

**Table 4** Performance of CNN architectures used for regression modeling.

| Number | Architecture | Parameters (millions) | Depth (layers) | PLCC | SROCC |
|---|---|---|---|---|---|
| 1 | AlexNet | 61.0 | 8 | 0.5010 | 0.4917 |
| 2 | DarkNet-53 | 41.6 | 53 | 0.6431 | 0.6442 |
| 3 | DenseNet201 | 20.0 | 201 | 0.7432 | 0.7195 |
| 4 | EfficientNet-B0 | 5.3 | 82 | 0.6656 | 0.6597 |
| 5 | Inception-ResNet-V2 | 55.9 | 164 | 0.7933 | 0.7867 |
| 6 | Inception-V3 | 23.9 | 48 | 0.6819 | 0.6780 |
| 7 | MobileNet-V2 | 3.5 | 53 | 0.5950 | 0.5304 |
| 8 | NASNet | 88.9 | + | 0.8098 | 0.7922 |
| 9 | NASNet-Mobile | 5.3 | + | 0.4761 | 0.4460 |
| 10 | ResNet101 | 41.6 | 101 | 0.7246 | 0.6933 |
| 11 | ResNet50 | 25.6 | 50 | 0.5862 | 0.5641 |
| 12 | Vgg16 | 138.0 | 16 | 0.5217 | 0.5193 |
| 13 | Xception | 22.9 | 71 | 0.6525 | 0.5770 |

Regularization is important because it prevents the model from overfitting. Because some of the modeling architectures (NASNet or Inception-ResNet-V2) have a large number of parameters, they can easily overfit the training data. The first regularization measure is image augmentation, which increases the effective dataset size while decreasing overfitting. For input image selection, random cropping is used, which results in a regularization measure and provides a consistent input image size. Furthermore, dropouts of 50% and 25% are used with fully connected layers to prevent overfitting.

## 5.4 *Choice of Loss Function*

Neural networks are trained usually through stochastic gradient descent-based optimizers and required a loss function for learning. Loss function acts as an objective function and used to evaluate a candidate solution, the algorithm tries to minimize it to reduce the loss. The selection of a loss function is important but challenging. In case of classification, we have chosen categorical cross-entropy because it is the most widely used loss function for classification and perform reasonably well. In case of regression, mean squared error (MSE) is the most commonly used cost function. Mean absolute error (MAE) and Huber loss are among the other popular loss functions, which are reported to perform better by some researchers.[11,39] It is reported that MAE is less sensitive to outliers, whereas the Huber is considered to be a balanced loss function between MSE and MAE.[40] To select a suitable loss function, we used ResNet50 as base architecture and trained it on BIQ2021 for the three loss functions that are defined as

$$\text{MSE} = \frac{1}{n}\sum_{i=1}^{n}(\text{MOS-predictions})^2, \tag{3}$$

$$\text{MAE} = \frac{1}{n}\sum_{i=1}^{n}|\text{MOS-predictions}|, \tag{4}$$

$$\text{Huber} = \frac{(|\Delta| \le 1)}{2} + \left((|\Delta| > 1) \times |\Delta| - \frac{1}{2}\right), \tag{5}$$

**Table 5** Performance evaluation of existing image quality evaluator.

| Number | Method | PLCC | SROCC |
|---|---|---|---|
| 1 | Naturalness Image Quality Evaluator | 0.51127 | 0.483828 |
| 2 | Integrated Local-NIQE | 0.541263 | 0.506898 |
| 3 | Discriminable Image Pairs-Image Quality | 0.34478 | 0.326032 |
| 4 | Psychovisual-based Image Quality Evaluator | 0.220406 | 0.270546 |
| 5 | Blind Image Quality Index | 0.568412 | 0.511794 |
| 6 | Blind/Referenceless Image Spatial QUality Evaluator | 0.718858 | 0.641661 |
| 7 | Blind Image Integrity Notator using DCT Statistics | 0.439035 | 0.367351 |
| 8 | Codebook Representation for No-Reference Image Assessment | 0.782518 | 0.720473 |
| 9 | Distortion Identification-based Image Verity and Integrity Evaluation | 0.61372 | 0.597999 |
| 10 | High Order Statistics Aggregation | 0.746956 | 0.735253 |
| 11 | Blind Image Quality | 0.8098 | 0.7922 |

where

$$|\Delta| = \text{MOS-prediction}.$$

### 5.5 *Performance Evaluation Using Conventional Approaches*

The proposed image quality database is a blind image quality database as there is no-reference information. Therefore, only no-reference or BIQA algorithms can be trained or validated on this database. Few popular image quality evaluators are selected based on the availability of their source code and evaluated on BIQ2021 database. The results of these image quality evaluators are obtained by retraining and testing. A train–test split of 80% to 20% is provided with dataset, and the same split is used here for fair comparison (Table 5).

## 6 Conclusion

A large-scale database on naturally distorted images is proposed in this study. The data contain images with varying content and level of distortions. The subjective scoring of the images is performed using laboratory-based absolute category rating. The laboratory-based scoring ensures reliable quality ratings under the supervision of a guide. The image quality is represented by these images through MOS and the quality category obtained through majority voting. The conventional MOS can be used for modeling a regression algorithm for quality prediction, and the quality category can be used to modeling a classification algorithm. The quality category for training a classification model is provided due to its utility in determining the quality rating of an image. These labels can be used to decide the selection or rejection of an image for a particular application[8] without supplying any threshold. The modeling of quality assessment for regression and classification is performed by designing an architecture based on eight different CNN architectures. Each of these selected architectures falls into different category of CNN and therefore demonstrates the usefulness of each CNN type for quality assessment performance. It is observed that NASNet is the most suitable CNN architecture for prediction of absolute or continuous quality labels of digital images. A comparison with 10 existing approaches is made with the architecture based on NASNet named BIQ512 for prediction of image quality and best performance is obtained. The proposed database is believed to bridge the gap of representative database of images having natural distortions and reliable annotations.

## Acknowledgments

The author declares that there is no potential confilict of interest.

## Code, Data, and Materials Availability

The dataset described in this paper is provided with a Creative Commons license at: https://github.com/nisarahmedrana/BIQ2021. The trained model is uploaded at MATLAB file exchange and its link along with the MATLAB codes are also provided at the same repository.

**Nisar Ahmed** received his PhD in computer engineering from the University of Engineering and Technology, Lahore, Pakistan, in 2022, and his master of science degree in computer engineering from the University of Engineering and Technology, Lahore, Pakistan, in 2015. His research interests include computer vision, pattern recognition, and digital image processing.

**Shahzad Asif** received his PhD in informatics from the University of Edinburgh, United Kingdom, in 2012. He is working as a professor and a chairman in the Department of Computer Science, University of Engineering and Technology, Lahore, Pakistan.